\pdfoutput=1

\documentclass[11pt]{article}

\usepackage{EMNLP2022}

\usepackage{times}
\usepackage{latexsym}

\usepackage[T1]{fontenc}

\usepackage[utf8]{inputenc}

\usepackage{times}
\usepackage{latexsym}
\usepackage{algorithm, algorithmic}
\usepackage{amsmath} 
\usepackage{amssymb}
\usepackage{multirow}
\usepackage{makecell}
\usepackage{hhline}
\usepackage{xspace}
\usepackage{booktabs}
\usepackage{float}
\usepackage{graphicx}
\renewcommand{\UrlFont}{\ttfamily\small}
\usepackage{amssymb}
\usepackage{pifont}
\usepackage{bm}
\usepackage{enumitem}
\usepackage{diagbox}
\setlist[itemize]{leftmargin=*}
\setitemize[1]{itemsep=0pt,partopsep=0pt,parsep=\parskip,topsep=0pt}
\usepackage{microtype}

\title{Improving Radiology Summarization with Radiograph and Anatomy Prompts}

\author{%
Jinpeng Hu$^{\heartsuit}$, \hspace{0.2cm}
Zhihong Chen$^{\heartsuit}$, \hspace{0.2cm}
Yang Liu$^{\heartsuit}$ \hspace{0.2cm} \\
 \textbf{Xiang Wan}$^{\heartsuit\diamondsuit\dag}$, \hspace{0.2cm} \textbf{Tsung-Hui Chang}$^{\heartsuit\dag}$ \\
$^{\heartsuit}$Shenzhen Research Institute of Big Data, The Chinese University of Hong Kong, \\Shenzhen, Guangdong, China \hspace{0.2cm} \\
$^{\diamondsuit}$Pazhou Lab, Guangzhou, 510330, China \hspace{0.2cm} \\
\texttt{
\{jinpenghu, zhihongchen, yangliu5\}@link.cuhk.edu.cn} \\
\texttt{wanxiang@sribd.cn} \hspace{0.2cm}
\texttt{changtsunghui@cuhk.edu.cn}
}

\begin{document}
\maketitle
\begin{abstract}
The impression is crucial for the referring physicians to grasp key information since it is concluded from the findings and reasoning of radiologists.
To alleviate the workload of radiologists and reduce repetitive human labor in impression writing, many researchers have focused on automatic impression generation.
However, recent works on this task mainly summarize the corresponding findings and pay less attention to the radiology images.
In clinical, radiographs can provide more detailed valuable observations to enhance radiologists' impression writing, especially for complicated cases.
Besides, each sentence in findings usually focuses on single anatomy, so they only need to be matched to corresponding anatomical regions instead of the whole image, which is beneficial for textual and visual features alignment.
Therefore, we propose a novel anatomy-enhanced multimodal model to promote impression generation. 
In detail, we first construct a set of rules to extract anatomies and put these prompts into each sentence to highlight anatomy characteristics. 
Then, two separate encoders are applied to extract features from the radiograph and findings. 
Afterward, we utilize a contrastive learning module to align these two representations at the overall level and use a co-attention to fuse them at the sentence level with the help of anatomy-enhanced sentence representation. 
Finally, the decoder takes the fused information as the input to generate impressions. 
The experimental results on two benchmark datasets confirm the effectiveness of the proposed method, which achieves state-of-the-art results.

\end{abstract}

\section{Introduction}
A radiology report of an examination is used to describe normal and abnormal conditions with one medical image and two important text sections: findings and impression.
%
%
The findings section is a free-text description of a clinical radiograph (e.g., chest X-ray) and dictates observations from the medical image.
Meanwhile, the impression is a more concise statement about critical observations summarized from the findings, images and the inference from radiologists and provides some clinical suggestions, such that in practice, clinicians prefer to read the impression to locate the prominent observations and evaluate their differential diagnoses.
However, writing impressions is time-consuming and in high demand, which draws many researchers to focus on automatic impression generation (AIG) to alleviate the workload of radiologists \cite{attend,hu2021word,zhang2018learning,karn2022differentiable,macavaney2019ontology,zhang2020optimizing}.
%
%

\begin{figure}[t]
\centering
\includegraphics[width=0.49\textwidth, trim=0 0 -10 0]{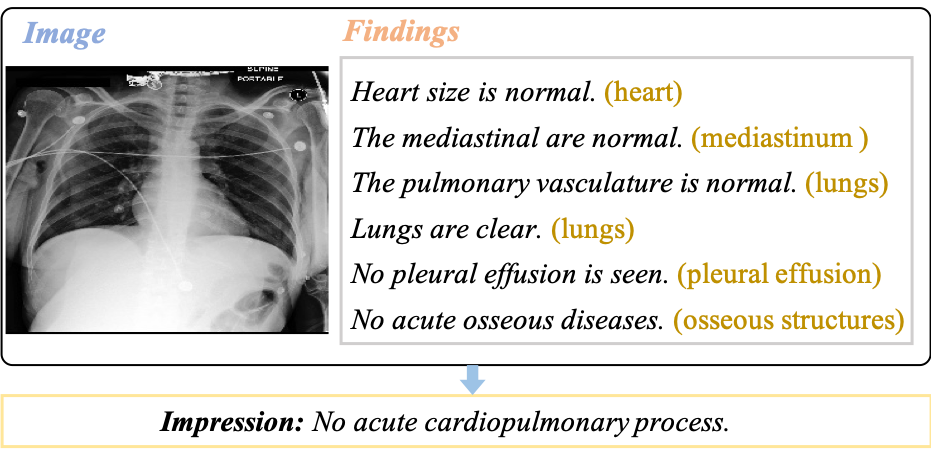}
\caption{An example of the radiology report and its chest X-ray image, where different color means that different sentences are aligned to the image.}
\label{fig:example_1}
\vskip -1em
\end{figure}

\begin{figure*}[t]
\centering
\includegraphics[width=0.995\textwidth, trim=0 0 0 20]{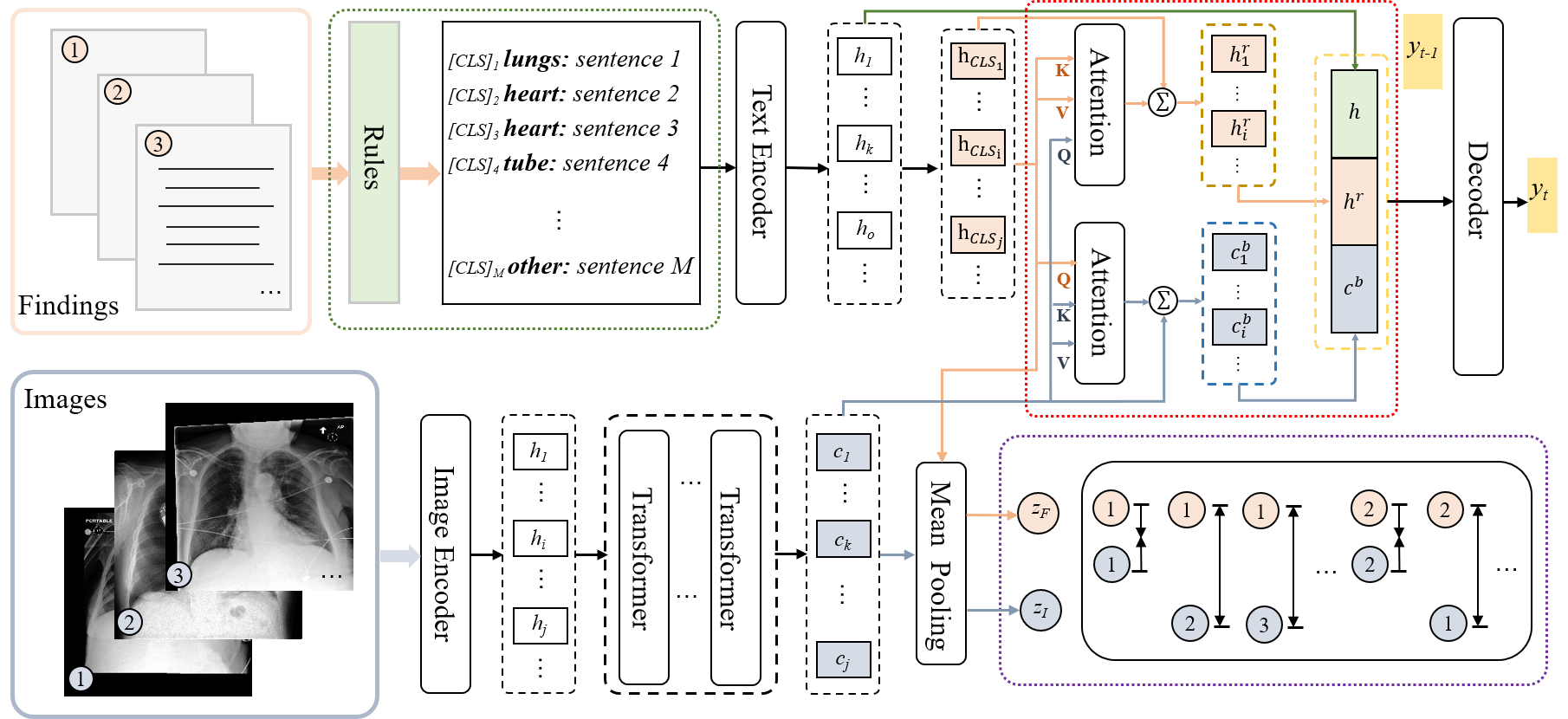}

\caption{
The overall architecture of our proposed model. 
The green box is used to provide sentence anatomy prompts. 
Besides, aligned contrastive learning and sentence-level co-attention fusion modules are shown in the purple and red boxes.
\textcircled{1}, \textcircled{2}, \textcircled{3} indicate different pairs (i.e., image and its corresponding findings).
}
\label{fig:architecture}
\vskip -1em
\end{figure*}

%
For example, \cite{attend,hu2021word,karn2022differentiable} propose to extract medical ontologies and entities from findings and then utilize graph neural networks (GNNs), dual encoder, or reinforcement learning to integrate this knowledge into general sequence-to-sequence models for promoting AIG.
Yet, most existing studies mainly focus on fully using findings to produce impressions and pay rare attention to medical radiography.
Owing to the fact that some diseases tend to have similar observations, they are difficult to get a clear diagnosis only depending on the textual statements.
In this situation, most radiologists usually consider both the image and findings to make a more accurate clinical suggestion in impressions.
Besides, many approaches have been proposed for radiology report generation and have achieved considerable success \cite{chen2021cross,zhang2020radiology}, whose goal is to generate the findings based on a given medical image, which further shows the value of knowledge in the medical image.
These report generation methods mainly follow a conventional encoder-decoder framework, where they employ a convolutional neural networks (CNNs) encoder to capture visual features and then incorporate these representations into a decoder to produce the reports.
In radiology reports, each findings can be regarded as a text representation of the corresponding medical image, and meanwhile, each image is a visual representation of the findings such that these two modal data can be effectively aligned.

Therefore, we propose a task that integrates the images and anatomy-enhanced findings for impression generation.
According to communication with radiologists, each sentence in the findings focuses on single anatomy, so the sentence-level representation should be easier to align to a certain anatomical region of the image.
To enhance such a process, we first construct some rules under the guidance of radiologists and then utilize such rules to extract the main anatomies from each sentence and put such anatomies at the beginning of the sentence to emphasize anatomy information.
Next, we utilize a visual extractor to extract visual features from the radiology image and apply a Transformer-based text encoder to embed the corresponding findings.
Afterward, an extra encoder is used to further model visual features, whose output will be aligned to the textual representation at the document level by a contrastive learning module.
Finally, we employ a co-attention to integrate the visual and text features at the sentence level to obtain the final fused representation, which is then input to the decoder to generate the impressions.
%
%
%
Experimental results on two benchmark datasets, MIMIC-CXR and OpenI, demonstrate the effectiveness of integrating images and findings.

\begin{table*}[ht]
\vspace{1mm}
\footnotesize
\centering
\resizebox{.99\textwidth}{!}{
\begin{tabular}{c|c}
\toprule
\multirow{1}{*} {\textbf{{Type}}} & \textbf{Keywords and Rules} \\
\midrule

\multirow{1}{*} {normal observations} & unremarkable, are normal, there are no, no ... seen, no ... present, ... \\
\multirow{1}{*} {lungs} & lung, lungs, pulmonary, suprahilar, perihilar, atelectasis, bibasilar, pneumonia, ... \\
\multirow{1}{*} {pleural spaces} & pleural \\
\multirow{1}{*} {heart} & heart, hearts, pericardial, cardiac, cardiopulmonary, cardiomediastinal, ... \\
\multirow{1}{*} {mediastinum} & mediastinal, mediastinum \\
\multirow{1}{*} {osseous structures} & fracture, osseous, glenohumeral, thoracic, bone, bony  \\
\multirow{1}{*} {tube} & tube, catheter  \\
\multirow{1}{*} {comparisons} & comparison, previous, prior  \\

\bottomrule
\end{tabular}}
 \vskip -0.5em
\caption{The details of the lexicon, where the left is the anatomy type and the right is the keywords and rules used to match the sentence.} 

  \label{Tab:lexicon}
\vskip -1em
\end{table*}

\section{Method}

We follow existing studies on report generation \cite{chen2020generating, zhou2021visual} and impression generation \cite{zhang2018learning, attend, hu2021word} and utilize the standard sequence-to-sequence paradigm for this task.
In doing so, we regard patch features extracted from radiology image $\mathcal{X_{I}}$ as one of the source inputs.
In addition, the other input is the findings sequence $\mathcal{X_{F}} = {s}_{1},{s}_{2},\cdots, {s}_{M}$, where $M$ is the number of sentence and ${s}_{i} = [CLS]_{i}, x_{i,1},x_{i,2},\cdots, x_{i,N_{i}}, [SEP]_{i}$ with external $[CLS]$ token.
The goal is to utilize $\mathcal{X_{I}}$ and $\mathcal{X_{F}}$ to find a target impression $\mathbf{Y}=\left[y_{1},...y_{i},..., y_{L}\right]$ that summarizes the most critical observations, where $L$ is the number of tokens and $y_{i} \in \mathrm{V}$ is the generated token and $\mathrm{V}$ is the vocabulary of all possible tokens. The impression generation process can be defined as:
\begin{equation}
\setlength\abovedisplayskip{6pt}
\setlength\belowdisplayskip{6pt}
    p(\mathbf{Y} \mid \mathcal{X_{I}}, \mathcal{X_{F}})=\prod_{t=1}^{L} p\left(y_{t} \mid y_{1}, \ldots, y_{t-1}, \mathcal{X_{I}}, \mathcal{X_{F}} \right).
\end{equation}
For this purpose, we train the proposed model to maximize the negative conditional log-likelihood of $\mathcal{Y}$ given the $\mathcal{X_{I}}$ and  $\mathcal{X_{F}}$:
\begin{equation}
\setlength\abovedisplayskip{6pt}
\setlength\belowdisplayskip{6pt}
    \theta^{*}=\underset{\theta}{\arg \max}  \sum_{t=1}^{L} \log p\left(y_{t} \mid y_{1}, ...,  y_{t-1}, \mathcal{X_{I}}, \mathcal{X_{F}};\theta\right),
\end{equation}
where $\theta$ can be regarded as parameters of the model.
The overall architecture of the proposed model is shown in Figure \ref{fig:architecture}.

\subsection{Visual Extractor}
\label{image-encoder}
We utilize a pre-trained convolutional neural networks (CNN) (e.g., ResNet \cite{he2016deep}) to extract features from $\mathcal{X_{I}}$.
We follow \citet{chen2020generating} to decompose the image into multiple regions with equal size and then expand these patch features into a sequence:
\begin{equation}
\setlength\abovedisplayskip{5pt}
\setlength\belowdisplayskip{5pt}
    [\mathbf{im}_{1},\mathbf{im}_{2},\cdots,\mathbf{im}_{P}] = {f}_{ve}(\mathcal{X_{I}}),
\end{equation}
where ${f}_{ve}$ refers to the visual extractor.

\subsection{Sentence Anatomy Prompts}
\label{sec:planning}
It is known that each sentence in findings usually focuses on describing observations in single anatomies, such as lung, heart, etc., instead of stating multiple anatomy observations in one sentence.
This might be because many radiologists usually draw on radiology report templates when writing findings, and most templates follow this characteristic, which describes medical observations anatomy by anatomy.
%
%
For example, radiology report templates in the radreport website\footnote{\UrlFont{https://radreport.org/}} mainly divide the radiology findings into six sections: Lungs, Pleural Spaces, Heart, Mediastinum, Osseous Structures, and Additional Findings, respectively.
Motivated by this, we manually construct a rule lexicon to extract anatomy information from the sentence, with the details shown in Table \ref{Tab:lexicon}.
After that, we utilize the following ways to deal with different types of sentences:
\begin{itemize}[leftmargin=*]
\setlength{\topsep}{0pt}
\setlength{\itemsep}{0pt}
\setlength{\parsep}{0pt}
\setlength{\parskip}{0pt}
    \item \textbf{Type} \uppercase\expandafter{\romannumeral1}: For the sentence that only describes observation in single anatomy, we assign the sentence to the corresponding anatomy type. For example, the sentence ``The lungs are hyperexpanded and mild interstitial opacities" only contains one anatomy (i.e., lungs), and thus, we assign type \textbf{lungs} to this sentence.

    \item \textbf{Type} \uppercase\expandafter{\romannumeral2}: Although most sentences focus on single anatomy, there are still some with multiple anatomies. For these sentences, we follow the priority ranking from \textbf{normal observations} to \textbf{comparisons}, as shown in Table \ref{Tab:lexicon}. For instance, although both \textbf{lung} and \textbf{pleural spaces} are in the sentence ``lungs are grossly clear, and there are no pleural effusions", we distribute this sentence into type \textbf{normal observations}.
    \item \textbf{Type} \uppercase\expandafter{\romannumeral3}: For the remaining sentences, we use a particular type \textbf{other observations} to mark. 
  
\end{itemize}
Next, we plan anatomy type into the corresponding sentence and modify the original sentence as ``anatomy: sentence".
For instance, the type \textbf{lungs} is inserted into ``The lungs are hyperexpanded and mild interstitial opacities" as ``lungs: The lungs are hyperexpanded and mild interstitial opacities".
In this way, the original findings $\mathcal{X}_{F}$ is updated as an anatomy-enhanced one $\mathcal{X}_{F}^{\prime}$.

\subsection{Text Encoder}
\label{text-encoder}
%
%
%
We employ a pre-trained model BioBERT \cite{BioBERT} as our text encoder to extract features from the findings:
\begin{equation}
\setlength\abovedisplayskip{5pt}
\setlength\belowdisplayskip{5pt}
    [\mathbf{h}_{1},\mathbf{h}_{2},\cdots,\mathbf{h}_{n}] = {f}_{te}(\mathcal{X}_{F}^{\prime}),
\end{equation}
where ${f}_{te}(\cdot)$ refers to the text encoder, and $\mathbf{h}_{i}$ is a high dimensional vector for representing tokens $x_{i}$. 
We regard the representation of $[CLS]_{i}$ in ${s}_{i}$ (i.e., $\mathbf{h}_{{CLS}_{i}}$) as the $i$th sentence representation.

\subsection{Document-level Cross-Modal Alignment}
In radiology reports, findings and radiology images usually describe the same medical observations by using different media (i.e., vision and text, respectively).
To pull the image representation close to the output of the text encoder, we first utilize an extra Transformer encoder to further model the visual features $\mathcal{X_{I}}$, computed by:
\begin{equation}
\setlength\abovedisplayskip{5pt}
\setlength\belowdisplayskip{5pt}
    [\mathbf{c}_{1},\mathbf{c}_{2},\cdots,\mathbf{c}_{P}] = {f}_{ie}(\mathbf{im}).
\end{equation}
Herein the outputs are the hidden states $\mathbf{c}_{i}$ encoded from the input visual features in,\ref{image-encoder} and $ {f}_{ie}$ refers to the Transformer image encoder.
Afterward, we utilize the mean pooling to obtain the overall representation with respect to the findings and the corresponding image, formalized as:
\begin{align}
\setlength\abovedisplayskip{6pt}
\setlength\belowdisplayskip{6pt}
\begin{split}
    \mathbf{z}_{I} = & Mean(\mathbf{c}_{1},\mathbf{c}_{2},\cdots,\mathbf{c}_{P}), \\
    \mathbf{z}_{F} = & Mean(\mathbf{h}_{{CLS}_{1}},\mathbf{h}_{{CLS}_{2}}, \cdots, \mathbf{h}_{{CLS}_{i}}).
\end{split}
\end{align}
Owing to the characteristic of the radiology report, $\mathbf{z}_{I}$ and $\mathbf{z}_{F}$ should be close to each other if the image and findings are from the same examination.
On the contrary, radiology images and reports from different tests tend to have distinct medical observations and further should be different from each other.
Therefore, we introduce a contrastive learning module to map positive samples closer and push apart negative ones, where the positive indicates that $\mathbf{z}_{I}$ and $\mathbf{z}_{F}$ are from the same pair (i.e., the same examination) with its report and radiology image and the negative refers to the samples from different pairs.
For example, we assume there are two tests, $(findings_1, images_1)$ and $(findings_2, image_2)$, and thus, in this case, for $findings_1$, the $image_1^{+}$ is a positive sample while the $image_2^{-}$ is a negative instance.
We follow \cite{gao2021simcse} to compute the cosine similarity between the original representation and its positive and negative examples.
Then, for a batch of $2Q$ examples $\mathbf{z} \in \{\mathbf{z}_{I}\} \cup \{\mathbf{z}_{F}\}$, we compute the contrastive loss for each $\mathbf{z}_{m}$ as:
\begin{equation}
\setlength\abovedisplayskip{5pt}
\setlength\belowdisplayskip{5pt}
    \mathcal{L}^{con}_{m} = -\log \frac{e^{\operatorname{sim}(\mathbf{z}_{m}, \mathbf{z}_{m}^{+}) /\tau}}
    {\sum_{\mathbf{z}^{-} \in \{\hat{\mathbf{z}}\}}(e^{\operatorname{sim}(\mathbf{z}, \mathbf{z}^{-}) / \tau})},
\end{equation}
where $\operatorname{sim}(\cdot,\cdot)$ is the cosine similarity, and $\tau$ is a temperature hyperparameter.
The total contrastive loss is the mean loss of all examples:
\begin{equation}
\setlength\abovedisplayskip{5pt}
\setlength\belowdisplayskip{5pt}
    \mathcal{L}^{con}=\frac{1}{2Q} \sum_{m=1}^{2Q} \mathcal{L}^{con}_{m}.
\end{equation}

\begin{table*}[t]
\footnotesize
\centering
\resizebox{.94\textwidth}{!}{
\begin{tabular}{l|p{3.5cm}|ccc|ccc}
\toprule[1pt]
\multirow{2}{*}{\textsc{\textbf{\makecell[c]{Data}}}} & \multirow{2}{*}{\textsc{\textbf{\makecell[c]{Model}}}} 
& \multicolumn{3}{c|}{\textsc{\textbf{ROUGE}}} & \multicolumn{3}{c}{\textsc{\textbf{FC}}}  \\  
& & \textsc{R-1}  & \textsc{R-2}  & \textsc{R-L}   &\textsc{\textbf{P}} & \textsc{\textbf{R}} &\textsc{\textbf{F-1}} \\
\midrule                       
\multirow{6}{*} {\makecell*[l]{\textsc{OpenI}}}

& \textsc{Base-Image} & {47.07} & {33.10} & {47.05} & {-} & {-} & {-} \\

& \textsc{Base-Finding} & {66.37} & {58.01} & {66.27} & {-} & {-} & {-} \\

& \textsc{Base} & {66.94} & {58.87} & {66.89} & {-} & {-} & {-} \\

\cmidrule(l){2-8}

& \textsc{Base+DCA} & {67.48} & {59.05} & {67.34} & {-} & {-} & {-} \\

& \textsc{Base+AP} & {67.66} & {58.89} & {67.51} & {-} & {-} & {-} \\

& \textsc{Base+AP+DCA} & \textbf{68.00} & \textbf{59.89}  & \textbf{67.87} 

& {-} & {-} & {-} \\

\midrule

\multirow{6}{*} {\makecell*[l]{\textsc{MIMIC-CXR}}}

& \textsc{Base-Image} &   {24.97} & {14.11} & {24.42} & {34.74} & {33.20} & {32.87} \\

& \textsc{Base-Finding} & {46.48} & {31.38} & {45.13} & {56.29} & {50.88} & {52.51} \\

& \textsc{Base} &         {46.54} & {31.32} & {45.09} & {57.51} & {51.45} & {52.93} \\

\cmidrule(l){2-8}

& \textsc{Base+DCA} &      {46.83} & {31.40} & {45.33} & {56.41} & {51.87} & {53.39} \\

& \textsc{Base+AP} &      {47.06} & {31.66} & {45.74} & {57.68} & {50.79} & {53.07} \\

& \textsc{Base+AP+DCA} & \textbf{47.63} & \textbf{32.03}  & \textbf{46.13}
& \textbf{58.91} & \textbf{53.22} & \textbf{54.55} \\

\bottomrule
 \end{tabular}
 }
\vskip -2mm
  \caption{The performance of all baselines and our model on test sets of \textsc{OpenI} and \textsc{MIMIC-CXR} datasets. \textsc{R-1}, \textsc{R-2} and \textsc{R-L} refer to ROUGE-1, ROUGE-2 and ROUGE-L. \textsc{P}, \textsc{R} and \textsc{F-1} represent precision, recall, and F1 score.}%
  \label{Tab:performance_on_different_base}
\vskip -4mm
\end{table*}

\subsection{Sentence-Level Co-Attention Fusion}
As mentioned in \ref{sec:planning}, each sentence in the findings usually focuses on single anatomy, meaning that sentence-level textual information can be mapped to corresponding anatomy regions in images.
Therefore, we propose to utilize the anatomy-enhanced sentence representation to align with the image.
In detail, as introduced in \ref{text-encoder}, we extract anatomy-enhanced sentence representations from the text encoder $\mathbf{h}_{CLS}=[\mathbf{h}_{{CLS}_{1}},\mathbf{h}_{{CLS}_{2}}, \cdots, \mathbf{h}_{{CLS}_{i}}]$, which are then used to perform co-attention to fuse two modal knowledge.
We first treat $\mathbf{h}_{CLS}$ as query and the corresponding image representations $\mathbf{c}$ as key and value matrix and compute the attention weight with the softmax function:
\begin{equation}
\setlength\abovedisplayskip{4pt}
\setlength\belowdisplayskip{4pt}
    \mathbf{a}^{b}_{i} = \text{Softmax}(\mathbf{h}_{{CLS}_{i}} \mathbf{c}^\mathrm{T}),
\end{equation}
where $\mathbf{a}^{b}_{i}$ can be viewed as a probability distribution over the image features, which is then used to compute a weighted sum:
\begin{equation}
\setlength\abovedisplayskip{4pt}
\setlength\belowdisplayskip{4pt}
    \mathbf{c}^{b}_{i}=\sum_{k} a_{i,k}^{b} \mathbf{c}_{k}.
\end{equation}
Afterward, on the contrary, $\mathbf{c}$ is regarded as the key and value matrix, and $\mathbf{h}_{CLS}$ is represented as the query.
We then adopt a similar method to obtain another fusion representation:
\begin{equation}
\setlength\abovedisplayskip{4pt}
\setlength\belowdisplayskip{4pt}
    \mathbf{h}^{r}_{i}=\sum_{k} a_{i,k}^{r} \mathbf{h}_{{CLS}_{k}}, \mathbf{a}^{r}_{i} = \text{Softmax}(\mathbf{c_{i}} \mathbf{h}_{CLS}).
\end{equation}
After that, we obtain the updated image and sentence representation by adding the fusion vectors to the original ones:
\begin{equation}
\setlength\abovedisplayskip{5pt}
\setlength\belowdisplayskip{5pt}
    \mathbf{c} = \mathbf{c} + \mathbf{c}^{b}, \mathbf{h}_{CLS} = \mathbf{h}_{CLS} + \mathbf{h}^{r}.
\end{equation}

\subsection{Decoder}
The backbone decoder in our model is the one from the standard Transformer, where $\mathbf{e}=[\mathbf{c}, \mathbf{h}_{CLS}, \mathbf{h}]$ is functionalized as the input of the decoder so as to improve the decoding process.
Then, the decoding process at time step $t$ can be formulated as the function of a combination of previous output (i.e., $y_{1},\cdots,y_{t-1}$) and the feature input (i.e., $\mathbf{e}$):
\begin{equation}
\setlength\abovedisplayskip{5pt}
\setlength\belowdisplayskip{5pt}
    y_{t} = f_{de}(\mathbf{e},y_{1},\cdots,y_{t-1}),
\end{equation}
where $f_{de}(\cdot)$ refers to the Transformer-based decoder, and this process will generate a complete impression.
We define the overall loss function as the linear combination of impression generation loss and contrastive objectives:
\begin{equation}
\setlength\abovedisplayskip{5pt}
\setlength\belowdisplayskip{5pt}
    \mathcal{L} = \mathcal{L}^{generator}+ \lambda \mathcal{L}^{con},
\end{equation}
where $\lambda$ is the tuned hyper-parameter controlling the weight of the contrastive loss.

\section{Experimental Setting}

\subsection{Dataset}
Our experiments are conducted on two benchmark datasets: OpenI \cite{demner2016preparing} and MIMIC-CXR \cite{johnson2019mimic}, respectively, which are described as follows:
\vskip -1em
\begin{itemize}[leftmargin=*]
    \setlength{\topsep}{0pt}
    \setlength{\itemsep}{0pt}
    \setlength{\parsep}{0pt}
    \setlength{\parskip}{0pt}
    \item \textbf{\textsc{OpenI}}: it is a public dataset containing 7,470 chest X-ray images and 3,955 corresponding reports collected by Indiana University.
    \item \textbf{\textsc{MIMIC-CXR}}: it is a large-scale radiography dataset with 473,057 chest X-ray images and 206,563 report.
\end{itemize}
We follow \citet{hu2021word} to remove the following cases: (a) incomplete reports without findings or impressions; (b) reports whose findings have fewer than ten words or impression has fewer than two words.
Besides, since some reports have multiple radiology images from different views, such as posteroanterior, anteroposterior and lateral, we only select one image from posteroanterior or anteroposterior.
As for partition, we follow \citet{chen2020generating} to split OpenI and MIMIC-CXR, where the former is split as 70\%/10\%/20\% for train/validation/test, and the latter follows its official split.

\begin{table*}[ht]
\small
\centering
\resizebox{.97\textwidth}{!}{
\begin{tabular}{l|rrr|rrr}
\toprule
{\multirow{2}{*}
{ \textsc{\textbf{\makecell[l]{\\Model}}}}}  & \multicolumn{3}{c|}{\textsc{\textbf{OpenI}}} & \multicolumn{3}{c}{\textsc{\textbf{MIMIC-CXR}}} \\
\cmidrule(r){2-7}
& \textsc{R-1}  & \textsc{R-2}  & \textsc{R-L}   & \textsc{R-1}        & \textsc{R-2}       & \multicolumn{1}{c}{\textsc{R-L}}  \\
\cmidrule(lr){1-7}
\textsc{R2Gen} \cite{chen2020generating}   & 50.68    & 38.02    & 50.62   & 24.68  & 14.45    & 24.12    \\
\textsc{R2GenCMN} \cite{chen2021cross}   & 51.30    & 34.35   & 51.27  & 24.73    & 14.04    & 24.25     \\

\cmidrule(lr){1-7}
\textsc{TransAbs} \cite{liu2019text}  & 62.90   & 53.51   &  62.71  & 46.17    & 29.06    & 43.86     \\
\textsc{WGSum} \cite{hu2021word} & 63.90    & 54.49    & 63.89   & 46.83   & 30.42    & 45.02    \\
\cmidrule(lr){1-7}
\textsc{CLIPAbs} \cite{radford2021learning} & 53.13    & 39.69    & 52.99   & 38.23   & 23.44    & 36.62    \\

\cmidrule(lr){1-7}
\textsc{Ours} & \textbf{68.00} & \textbf{59.89}  & \textbf{67.87}  &  \textbf{47.63} & \textbf{32.03}  & \textbf{46.13} \\
\bottomrule
\end{tabular}}
\linespread{1}
\caption{Comparisons of our proposed models with the previous studies on the test sets of \textsc{OpenI} and \textsc{MIMIC-CXR} with respect to the ROUGE metric. 
More details on these existing studies are in \ref{appendix:baselinesettings}.
}
\label{Tab:performance_on_different_model}
\vskip -1em
\end{table*}

\subsection{Baseline and Evaluation Metrics}
To illustrate the validity of our proposed model, we utilize the following models as our main baselines:
\vskip -1em
\begin{itemize}[leftmargin=*]
    \setlength{\topsep}{0pt}
    \setlength{\itemsep}{0pt}
    \setlength{\parsep}{0pt}
    \setlength{\parskip}{0pt}
    \item \textbf{\textsc{Base-Findings}} and \textbf{\textsc{Base-Image}}: They are unimodal models, where the former utilizes a pre-trained text encoder and a randomly initialized Transformer-based decoder, and the latter replaces the text encoder with image encoders.
    
    \item \textbf{\textsc{Base}}: This is the base backbone multimodal summarization model with pre-trained image and text encoders and a Transformer-based decoder, which utilizes both findings and images to generate impressions.

    \item \textbf{\textsc{Base+DCA}} and \textbf{\textsc{Base+AP}}: They are the multimodal summarization models. The former utilizes document-level representations to align findings and images, and the latter utilizes the rules to enhance anatomy prompts for each sentence to promote feature fusion.

\end{itemize}
\noindent

We follow \citet{zhang2020optimizing} to utilize summarization and factual consistency (FC) metrics to examine the model performance. 
Specially, we utilize ROUGE \cite{lin2004rouge} and report $\textsc{F}_{1}$ scores of ROUGE-1 (R-1), ROUGE-2 (R-2), and ROUGE-L (R-L) for summarization metrics\footnote{For convenience, we utilize the python-implemented libraries to calculate the Rouge scores, which can obtain from \url{https://github.com/pltrdy/rouge}, which might be slightly different from the official ROUGE script.}.
Meanwhile, a pre-trained CheXbert \cite{smit2020chexbert}\footnote{We follow \citet{hu2021word} to evaluate the FC of MIMIC-CXR, which can be obtained from \url{https://github.com/stanfordmlgroup/CheXbert}} is used to recognize 14 types of observation from reference and generated impression, respectively, whose detected results are used to calculate the precision, recall, and F1 score for measuring FC.

\subsection{Implementation Details}
In our experiments, we utilize biobert-base-cased-v1.1\footnote{\url{https://github.com/dmis-lab/biobert}} as our text encoder and follow its default model settings which are 12 layers of self-attention with 768-dimensional embeddings.
Besides, for the visual extractor, we select the ResNet101 pre-trained on the ImageNet to extract patch features with the dimension 2048.
For the Transformer image encoder, we use a 6-layer Transformer with 768 hidden sizes and 2048 feed-forward filter sizes.
The decoder has a similar structure: 6-layer Transformer with 768 dimensions, 8 attention heads, and 2048 feed-forward filter sizes.
As for training, we use Adam \cite{kingma2014adam} to optimize the trainable parameters in our model, including the ones in pre-trained image and text encoders.

\section{Experimental Results}

\subsection{Overall Results}
To explore the effect of integrating image and text to generate impressions, we compare our model to corresponding single modal summarization baselines in Table \ref{Tab:performance_on_different_base} .
We can observe that compared to \textsc{Base-Findings} and \textsc{Base-Image}, all other models (except \textsc{Base}) obtain better results with respect to ROUGE scores, which shows the value of multimodal information fusion.
The main reason might be that findings can provide key and accurate information, and the image can present detailed and rich features, such that these two different types of features can complement each other to promote impression generation.
Besides, \textsc{Base-Findings} outperforms \textsc{Base-Image}, illustrating that textual features are more valuable than visual ones in this task because the gap between two related texts is smaller than that between vision and text.

Moreover, we conduct experiments on the different models, and the results are reported in Table \ref{Tab:performance_on_different_base} where \textsc{Base+AP+DCA} indicates our full model.
There are several observations drawn from different aspects.
First, the comparisons between \textsc{Base+DCA}, \textsc{Base+AP}, and \textsc{Base} illustrate the effectiveness of each component in our proposed model (i.e., contrastive learning and lexicon matching).
Second, our full model (i.e., \textsc{Base+AP+DCA}) achieves the best results among these baselines, which confirms the validity of our design that combines contrastive learning and anatomy information planning.
Contrastive learning can map the image closer to the corresponding findings if they are in the same pair and push them apart if they are not, which can effectively align these two multimodal features at the document level.
For another, highlighting anatomy characteristics can potentially help the model align the sentence feature to the corresponding organ or body part position in the images, further improving feature fusion between different multimodal.
Third, in terms of FC metrics on the MIMIC-CXR dataset, our proposed model outperforms all baselines and achieves higher F1 scores, indicating that our model is able to generate more accurate impressions.
This is because our model can enhance feature matching between visual and textual features by aligning and fusion to promote critical information extraction, contributing to better impression generation with the help of such information.

\subsection{Comparison with Previous Studies}
We further compare our model with existing methods, with the results reported in Table \ref{Tab:performance_on_different_model}.
We can observe that our model achieves the best results among these methods, although those studies utilize complicated structures to enhance the generation, e.g., \textsc{WGSum} utilizes a complicated graph structure, and \textsc{R2Gen} uses a recurrent relational memory.
In addition, it is surprise that \textsc{CLIPAbs} achieve worse performance than \textsc{TransAbs} and \textsc{WGSum}.
This might be because CLIP \cite{radford2021learning} pays more attention to the images and is less powerful in encoding text, while textual features are more important in this task.

\begin{table}[t]
\centering
\resizebox{.49\textwidth}{!}{
\begin{tabular}{lcrrr}

\toprule[1pt]
{\textbf{Comparison}} & {\textbf{Metric}} &{\textbf{Win}} & {\textbf{Tie}} &{\textbf{Lose}} \\
\midrule
\multirow{3}{*} {\makecell*[l]{Ours vs. Base}} 

& \textsc{Read.} & {8\%} & {88\%}  & {4\%} \\
& \textsc{Acc.} & {25\%} & {58\%}  & {17\%}\\
& \textsc{Comp.} & {13\%} & {80\%}  & {7\%} \\

\cmidrule(r){1-5}
\multirow{3}{*} {\makecell*[l]{Ours vs. Ref}} 
& \textsc{Read.} & {4\%} & {77\%}  & {9\%} \\
& \textsc{Acc.} & {12\%} & {70\%}  & {18\%} \\
& \textsc{Comp.} & {5\%} & {85\%}  & {10\%} \\

\bottomrule
\end{tabular}}
\vskip -0.5em
\caption{Results of the human evaluation. The top three give results for comparison between \textsc{Base+AP+DCA} and \textsc{Base}. The bottom three are results for \textsc{Base+AP+DCA} versus the reference impressions.}

  \label{Tab:human}
\vskip -1em
\end{table}
\subsection{Human Evaluation}
We also conduct a human evaluation to evaluate the quality of the generated impressions with respect to three metrics: Readability, Accuracy, and Completeness \cite{attend}.
In detail, we randomly select 100 chest X-ray images and their findings and impressions from the test set of MIMIC-CXR, as well as impressions generated from different models.
Afterward, three experts who are familiar with radiology reports are invited to evaluate the generated impression with the results shown in Table \ref{Tab:human}.
We can observe that our model is better than \textsc{Base}, where more impressions from our model have higher quality than those from \textsc{Base}, further confirming the effectiveness of our model.
Meanwhile, when comparing our model against references, we find that although some cases are worse than ground truth (9\%, 18\%, and 10\%), most of the impressions from our model are at least as good as the reference impressions.


\section{Analyses}
\subsection{Impression Length}
\begin{figure}[t]
\centering
\includegraphics[width=0.5\textwidth, trim=0 0 0 45]{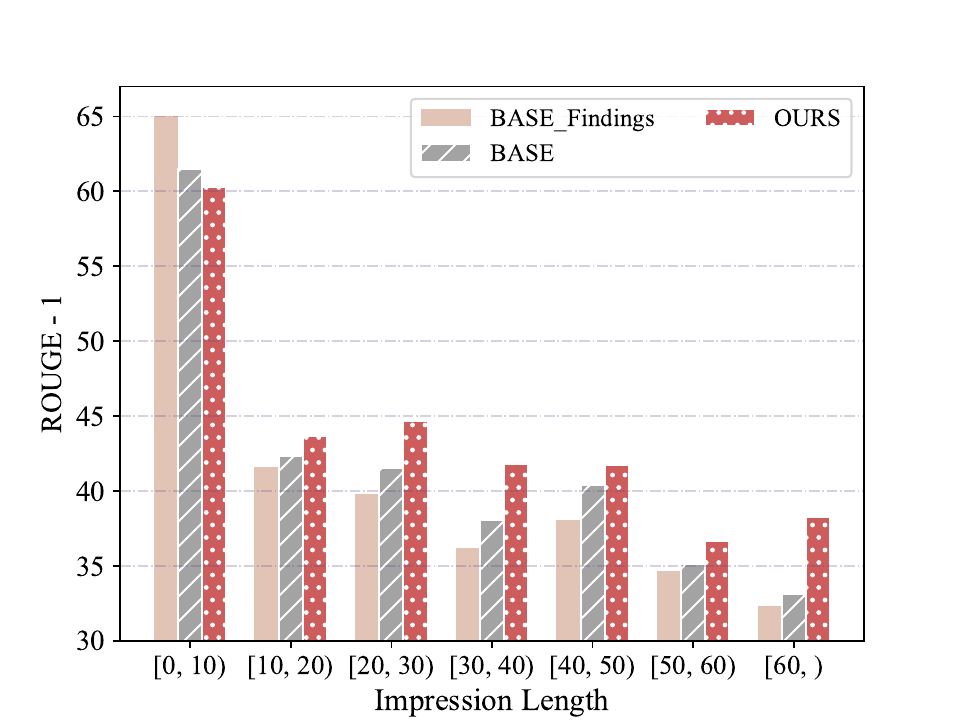}
\vskip -0.5em
\caption{R-1 score of generated impressions from different models, where OURS represents the \textsc{Base+AP+DCA}. Note that when the impression length is longer than 20, the $p$ is less than 0.05.}
\label{fig:effect-length}
\vskip -1em
\end{figure}

To test the effect of the length of impressions in AIG, we categorize the generated impressions on the MIMIC-CXR test set into several groups according to the length of reference impression, with the R-1 scores shown in Figure \ref{fig:effect-length}.
Note that the average impression length for MIMIC-CXR is 17.
We can observe that these models tend to have worse performance with the increasing impression length, especially in the last group, where all obtain the worst R-1 scores.
Our proposed model achieves more promising results in most groups, except the first group where the \textsc{Base-Findings} achieves the best results, which illustrates that our model is better at generating longer impressions.
The main reason is that short impressions are usually normal observations without complicated abnormalities so that findings are enough to describe such information, and images may lead to some redundant noise due to their being too detailed.
In contrast, for the long impression, this more detailed information can complement textual features to help the model accurately grasp complex observations.

\begin{figure*}[t]
\centering
\includegraphics[width=0.99\textwidth, trim=0 0 0 30]{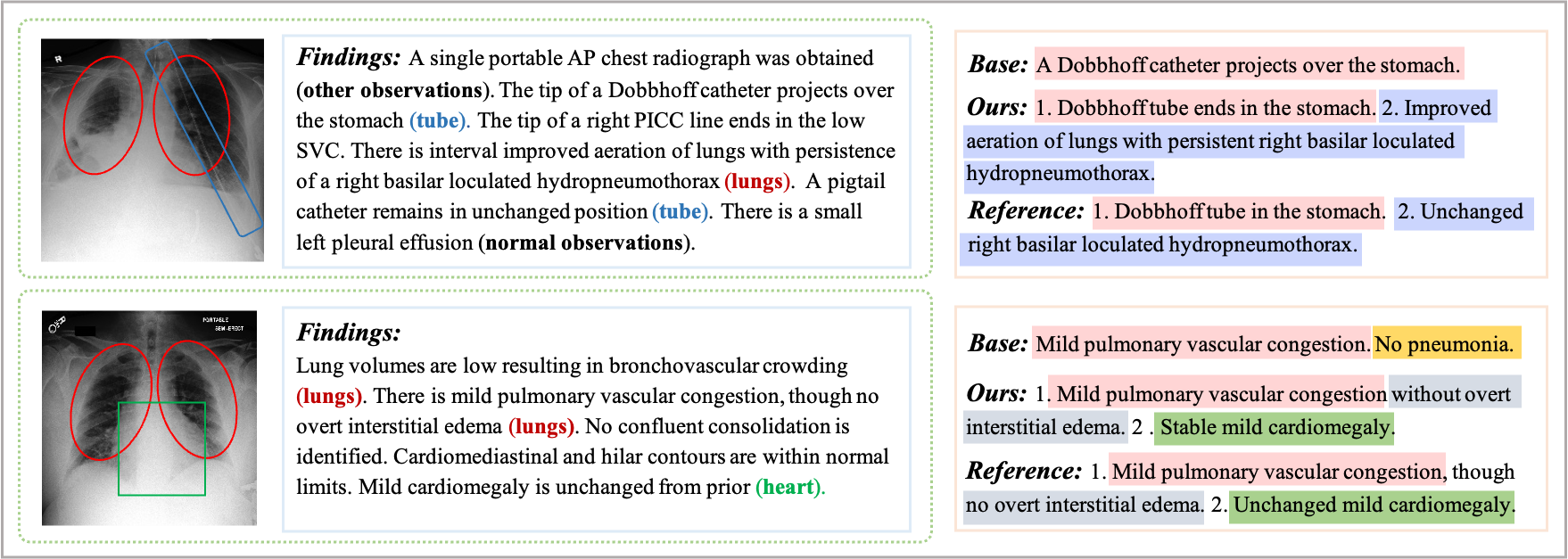}
\vskip -0.8em
\caption{Examples of the generated impressions from \textsc{Base} and \textsc{Base+AP+DCA} as well as reference impressions. Lungs, tubes and hearts are located in the red, blue and green boxes.}
\label{fig:case}
\vskip -1em
\end{figure*}

\subsection{Case Study}
To further qualitatively investigate the effectiveness of our proposed model, we conduct a case study on the generated impressions from different models whose inputs are X-ray images and corresponding findings.
The results are shown in Figure \ref{fig:case}, and different colors represent the observations found in different locations.
It is observed that \textsc{Ours} is able to produce better impressions than the \textsc{Base} model, where impressions from our models can almost cover all the key points in these two examples with the help of the corresponding regions in images.
On the contrary, the \textsc{Base} model ignores some critical observations written in reference impressions, such as \textit{``right basilar loculated hydropneumothorax.''} in the first example and \textit{``Stable mild cardiomegaly''} in the second example, and even generates some unrelated information (e.g., \textit{``No pneumonia''} in the second case).

\section{Related Work}
\subsection{Multimodal Summarization}
With the increase of multimedia data, multimodal summarization has recently become a hot topic, and many works have focused on this area, whose goal is to generate a summary from multimodal data, such as textual and visual \cite{zhu2018msmo, li2018multi,zhu2020multimodal,li2020multimodal2, im2021self,atri2021see,delbrouck2021qiai}.
For example, \citet{li2017multi} proposed to generate a textual summary from a set of asynchronous documents, images, audios and videos by a budgeted maximization of submodular functions.

\subsection{Radiology report generation}
Image captioning is a traditional task and has received extensive research interest \cite{you2016image,aneja2018convolutional,xu2021towards}.
Radiology report generation can be treated as an extension of image captioning tasks to the medical domain, aiming to describe radiology images in the text (i.e., findings), and has achieved considerable improvements in recent years \cite{chen2020generating,zhang2020radiology,liu2019clinically,liu2021exploring,zhou2021visual,boag2020baselines,pahwa2021medskip,jing2019show,zhang2020contrastive,you2021aligntransformer,liu2019aligning}.
\citet{liu2021competence} employed competence-based curriculum learning to promote report generation, which started from simple reports and then attempted to consume harder reports.

\subsection{Radiology impression generation}
Summarization is a fundamental text generation task in natural language processing (NLP), drawing sustained attention over the past decades \cite{see2017get,liu2019text,duan2019contrastive,chen2018fast,lebanoff2019scoring}.
General Impression generation can be regarded as a special type of summarization task in the medical domain, aiming to summarize findings and generate impressions.
There are many methods proposed for this area \cite{attend,hu2021word,zhang2018learning,hu2022graph,karn2022differentiable,macavaney2019ontology,zhang2020optimizing}.
\citet{macavaney2019ontology,attend} proposed to extract medical ontologies and then utilize a separate encoder to extract features from such critical words for improving the decoding process and thus promote AIG.
\citet{hu2021word} further constructed a word graph by medical entities and dependence tree and then utilized the GNN to extract features from such graph for guiding the generation process.
However, recent works on this area mainly focus on the text section, that is, using the findings to generate the impression, and thus they have not explored the valuable information in corresponding radiology images.

\section{Conclusion}
This paper proposes an anatomy-enhanced multimodal summarization framework to integrate radiology images and text for facilitating impression generation.
In detail, for radiology images, we utilize a visual extractor to extract detailed visual features.
For radiology findings, we first plan anatomical prompts into each sentence by keywords and rules and then utilize a pre-trained encoder to distillate features from modified findings.
Afterward, we employ a contrastive learning module to align the visual and textual features at the document level and use a co-attention to fuse these two features at the sentence level, which are then input to the decoder to improve impression generation.
Furthermore, experimental results on two benchmark datasets illustrate the effectiveness of our proposed model, which achieves considerable improvements.
%

\section{Limitations}
Although our model has achieved considerable improvements, as shown in Figure \ref{fig:effect-length}, our model tends to have a slight decrease in short impression generation, which need to be further solved in the future.
In this paper, we follow previous studies and only utilize English radiology report datasets to verify the effectiveness of our proposed model, which is limited in verification in other languages.
The main reason is that most publicly available radiology report datasets center on the English.
%

\bibliography{anthology,custom}
\bibliographystyle{acl_natbib}



\clearpage
\appendix
\section{Appendix}
\label{sec:appendix}

\subsection{Baseline Settings}
\label{appendix:baselinesettings}

We also compare our model with several existing studies.
For unimodal summarization models, we select several report generation models, e.g., \textbf{\textsc{R2Gen}} \cite{chen2020generating}, \textbf{\textsc{R2GenCMN}} \cite{chen2021cross} and several radiology summarization models, e.g., \textbf{\textsc{TransAbs}} \cite{liu2019text}, \textbf{\textsc{WGSum}} \cite{hu2021word}.
For multimodel summarization models, we compare our model with \textbf{\textsc{CLIPAbs}}, which utilizes CLIP \cite{radford2021learning} to extract multimodal features.

For \textsc{R2Gen}, \textsc{R2GenCMN} and \textsc{WGSum}, we utilize their released codes to replicates results.
For \textsc{CLIPAbs}, we utilize the clip-vit-base-patch32 as the encoder.
Notice that for all these baselines, we utilize the python-implement to calculate the ROUGE scores instead of the official ROUGE script.

\subsection{Dataset}
\label{Dataset_statistic}
\begin{table}[t]
\footnotesize
\centering
\resizebox{.46\textwidth}{!}{
\begin{tabular}{l|l|r|r|r}
\toprule[1pt]
{\textsc{\textbf{Data}}}&\textsc{\textbf{Type}}  &\textsc{\textbf{Train}} & \textsc{\textbf{Dev}} &\textsc{\textbf{Test}} \\
\midrule                       
\multirow{5}{*} {\makecell*[l]{\textsc{OpenI}}}
& \textsc{Report \#} & {2.4K} & {0.3K} & {0.6K}\\
& \textsc{Avg. wf} & {37.9} & {37.8}  & {30.0} \\
& \textsc{Avg. sf} & {5.75} & {5.68}  & {5.77} \\
& \textsc{Avg. wi} & {10.4} & {11.2}  & {10.6} \\
& \textsc{Avg. si} & {2.86} & {2.94}  & {2.82} \\
\midrule
\multirow{6}{*} {\makecell*[l]{\textsc{MIMIC} \\ {-CXR}}}
& \textsc{Report \#} & {117.7K} & {0.9K} & {1.5K}\\
& \textsc{Image \#} & {117.7K} & {0.9K} & {1.5K}\\
& \textsc{Avg. wf} & {55.4} & {56.3}  & {70.0} \\
& \textsc{Avg. sf} & {5.49} & {5.51}  & {6.24} \\
& \textsc{Avg. wi} & {16.4} & {16.26}  & {21.1} \\
& \textsc{Avg. si} & {1.66} & {1.65}  & {1.87} \\
\bottomrule
 \end{tabular}}
\vskip -0.3em
\linespread{1}
\caption{The statistics of the two benchmark datasets with random split for \textsc{OpenI} and official split for \textsc{MIMIC-CXR}, including the numbers of report, the averaged sentence-based length (\textsc{Avg. sf}, \textsc{Avg. si}), the averaged word-based length (\textsc{Avg. wf}, \textsc{Avg. wi}) of both \textsc{Impression} and \textsc{Findings}.}%
  \label{Tab:dataset}
\vskip -1em
\end{table}
We present the statistics of these two datasets in Table \ref{Tab:dataset}.

\end{document}